\documentclass{article}

     \usepackage[final,nonatbib]{neurips_2019}

\usepackage[utf8]{inputenc} % allow utf-8 input
\usepackage[T1]{fontenc}    % use 8-bit T1 fonts
\usepackage[hidelinks]{hyperref}
\hypersetup{
	colorlinks = true,
	citecolor=blue
}
\usepackage{url}            % simple URL typesetting
\usepackage{booktabs}       % professional-quality tables
\usepackage{amsfonts}       % blackboard math symbols
\usepackage{nicefrac}       % compact symbols for 1/2, etc.
\usepackage{microtype}      % microtypography
\usepackage{graphicx}
\usepackage{amsmath,amssymb,amsfonts}
\usepackage{subfig}
\usepackage{array}
\newcolumntype{P}[1]{>{\centering\arraybackslash}p{#1}}
\usepackage{comment}
\newcommand{\X}{\mathbb{X}}
\newcommand{\Y}{\mathbb{Y}}

\newcommand{\Xtri}{\mathbb X_{\mathrm{train}}^i}
\newcommand{\Ytri}{\mathbb Y_{\mathrm{train}}^i}

\usepackage{authblk}

\title{Differential Privacy-enabled Federated Learning for Sensitive Health Data}

\author[1, *]{\textbf{Olivia Choudhury}}
\author[2]{\textbf{Aris Gkoulalas-Divanis}}
\author[3]{\textbf{Theodoros Salonidis}}
\author[1]{\textbf{Issa Sylla}}
\author[1] {\\\textbf{Yoonyoung Park}}
\author[4, **]{\textbf{Grace Hsu}}
\author[1] {\textbf{Amar Das}}

\affil[1]{\footnotesize IBM Research Cambridge, Cambridge, MA, USA}
\affil[2]{\footnotesize IBM Watson, Cambridge, MA, USA}
\affil[3]{\footnotesize IBM T.J. Watson Research Center, Yorktown Heights, NY, USA}
\affil[4]{\footnotesize Massachusetts Institute of Technology, Cambridge, MA, USA}
\affil[*]{\footnotesize Corresponding author: \texttt{olivia.choudhury1@ibm.com}}
\affil[**]{\footnotesize Research done during internship at IBM Research Cambridge}

\begin{document}

\maketitle

\begin{abstract}
 
Leveraging real-world health data for machine learning tasks requires addressing many practical challenges, such as distributed data silos, privacy concerns with creating a centralized database from person-specific sensitive data, resource constraints for transferring and integrating data from multiple sites, and risk of a single point of failure. In this paper, we introduce a federated learning framework that can learn a global model from distributed health data held locally at different sites. The framework offers two levels of privacy protection. First, it does not move or share raw data across sites or with a centralized server during the model training process. Second, it uses a differential privacy mechanism to further protect the model from potential privacy attacks. We perform a comprehensive evaluation of our approach on two healthcare applications, using real-world electronic health data of 1 million patients. We demonstrate the feasibility and effectiveness of the federated learning framework in offering an elevated level of privacy and maintaining utility of the global model.

\end{abstract}

\section{Introduction}

Learning from real-world health data has proven effective in multiple healthcare applications, resulting in improved quality of care~\cite{gultepe2013vital, johnson2016machine, chakrabarty2019getting}, generating medical image diagnostic tools~\cite{erickson2017machine, menze2014multimodal, gibson2018niftynet}, predicting disease risk factors~\cite{poplin2018prediction, zheng2017machine, kourou2015machine}, and analyzing genomic data for personalized medicine~\cite{libbrecht2015machine, choudhury2018hecil, beam2018big}. Healthcare data is often segregated across data silos, which limits its potential for insightful analytics. Data access is further restricted due to regulatory policies mandated by the US Health Insurance Portability and Accountability Act (HIPAA)\footnote[1]{https://www.hhs.gov/hipaa/for-professionals/privacy/special-topics/de-identification/index.html} and EU General Data Protection Regulation (GDPR)~\cite{spainDPA}. Traditional or centralized machine learning algorithms require aggregating such distributed data into a central repository for the purpose of training a model. However, this incurs practical challenges, such as regulatory restrictions on sharing patient-level sensitive data, high resources required for transferring and aggregating the data, as well as a high risk associated with introducing a single point of failure. Leveraging such data while complying with data protection policies requires re-thinking data analytics methods for healthcare applications. 

Federated learning (FL) offers a new paradigm for training a global machine learning model from data distributed across multiple data silos, eliminating the need for raw data sharing~\cite{mcmahan2016communication}. Once a global model is shared across sites, each site trains the model based on its local data. The parameter updates of the local models are subsequently sent to an aggregation server and incorporated into the global model. This is repeated until a convergence criterion of the global model is satisfied. The merit of FL has been recently demonstrated in several real-world applications, including image classification~\cite{wang2019adaptive} and language modeling~\cite{mcmahan2016communication}. It is especially relevant in healthcare applications, where data is rife with personal, highly-sensitive information, and data analysis methods must comply with regulatory requirements~\cite{brisimi2018federated}. 

In this paper, we propose, implement, and evaluate a FL framework for analyzing distributed healthcare data. FL provides a first level of privacy protection by training a global model without sharing raw data among sites. However, in certain cases, FL may be vulnerable to inference attacks  ~\cite{bagdasaryan2018backdoor,bonawitz2017practical,geyer2017differentially}. Although state-of-the-art approaches to address such attacks are based on differential privacy~\cite{dwork2006our}, their performance has not been investigated for healthcare applications. To this end, we extend the FL framework with a distributed differential privacy mechanism and investigate its performance. 

We perform a comprehensive empirical evaluation using two real-world health datasets comprising electronic health records and administrative claims data of 1 million patients. We analyze the effect of different levels of privacy on the performance of the FL model and derive insights about the effectiveness of FL with and without differential privacy in healthcare applications. More specifically, we show that FL without differential privacy can provide model performance close to the hypothetical case where all data is centralized. We then study how $\epsilon$-differential privacy impacts the performance of the produced global FL model for a given level of privacy. The results show that although differential privacy offers a strong level of privacy, it deteriorates the predictive capability of the produced global models due to the excessive amount of noise added during the distributed FL training process.

\section{Methods}

\subsection{Federated learning}

For the purpose of illustration, we consider classification algorithms that are amenable to gradient descent optimization. For a general binary classification problem, let the features, denoted by $x_k$ (for the $k^{th}$ feature), be drawn from a feature space $\X$. The corresponding labels $y_k$ are drawn from the label space $\Y:= \{-1,1\}$. Let the features corresponding to positive labels be denoted by $\X_+$ and those corresponding to negative labels by $\X_-$, that is 
\[
\X_+ = \{x_k \in \X : y_k = +1 \} \quad \text{and} \quad \X_- = \{x_k \in \X : y_k = -1 \}.
\]
For any $x^+_k\in \X_+$ and $x^-_k\in \X_-$, the objective of this classification is to construct a function $f:\X\to\Y$ such that
\[
    f(x_k^+)= +1 \quad \text{and} \quad f(x_k^-) = -1.
\]

As discussed later, for our use cases of predicting adverse drug reaction and in-hospital mortality, we denote cases of ADR and in-hospital mortality as labels $y_k = +1$ , and cases of non-ADR and non-mortality as $y_k = -1$. For a FL setup with $N$ sites, let $\mathcal{D}_i$, with feature set $\{\Xtri\}_{i=1}^N$ and corresponding label set $\{\Ytri\}_{i=1}^N$  be the local training data at site $i$, where $i \in N$. Based on the use case, a global model is shared with each site, which trains the model on its local data $\mathcal{D}_i$, where $i \in N$. During local model training, based on the given learning rate, number of epochs, and batch, we compute average gradient ($\triangledown F_i(w)$) with respect to its current model parameter $w$. We then compute weighted average to aggregate the parameter updates from the local models. The process is repeated until a convergence criterion, such as minimization of loss function, is satisfied. For further details on implementing a FL model for healthcare applications, we refer the readers to~\cite{choudhuryAMIA}.

\subsection{Differential Privacy}

Differential privacy~\cite{dwork2006calibrating,dwork2011firm,dwork2014algorithmic} is a widely-used standard for privacy guarantee of algorithms operating on aggregated data. A randomized algorithm $\mathcal{A}(\mathcal{D})$ satisfies $\epsilon$-differential privacy if for all datasets $\mathcal{D}$ and $\mathcal{D'}$, that differ by a single record, and for all sets $\mathcal{S} \in \mathcal{R}$, where $\mathcal{R}$ is the range of $\mathcal{A}$,

\[
Pr[\mathcal{A}(\mathcal{D}) \in \mathcal{S}] \leq e^\epsilon Pr[\mathcal{A}(\mathcal{D'}) \in \mathcal{S}].
\]

where $\epsilon$, a privacy parameter, is a non-negative numbers. This implies that any single record in the dataset does not have a significant impact on the output distribution of the algorithm. 

There are several methods for generating an approximation of $\mathcal{A}$ that satisfies differential privacy. Based on the different approaches of adding noise, they can be categorized into input perturbation, output perturbation, exponential mechanism, and objective perturbation~\cite{sarwate2013signal}. Existing work on the application of differential privacy in machine learning have primarily focused on models that are trained on a centralized dataset~\cite{abadi2016deep,ji2014differential,sarwate2013signal}. As a recent advancement in privacy-preserving FL, the authors in~\cite{geyer2017differentially} adopted output perturbation or \textit{sensitivity method}. However, prior research has shown the effectiveness of objective perturbation, with theoretical guarantee, in outperforming the output perturbation approach~\cite{chaudhuri2011differentially}. Hence, in this work, we explore the potential of differential privacy, based on objective perturbation, in the context of FL. For the task of classification, we add noise to the objective function of the optimization to obtain a differentially private approximation. At each site, the noise is added to the objective function of the model to produce a minimizer of the perturbed objective. For a comprehensive background on differential privacy with objective perturbation, we refer the readers to~\cite{chaudhuri2009privacy,chaudhuri2011differentially}.

\section{Evaluation}

\subsection{Use cases and data preparation}

Developing FL models in a privacy-preserving manner is very important, especially in the context of healthcare, where patient data are extremely sensitive. To evaluate our proposed approach, we consider two major tasks for improving the health outcome of patients: (a) prediction of adverse drug reaction (ADR), and (b) prediction of mortality rate. ADR is a major cause of concern amongst medical practitioners, pharmaceutical industry, and the healthcare system\footnote[3]{https://www.fda.gov/drugs/informationondrugs/ucm135151.htm}. As healthcare data is distributed across data silos, obtaining a sufficiently large dataset to detect such rare events poses a challenge for centralized learning models. For the purpose of ADR prediction, we used Limited MarketScan Explorys Claims-EMR Data (LCED), which comprises administrative claims and electronic health records (EHRs) of over 1 million commercially insured patients. It consists of patient-level sensitive features, such as demographics, habits, diagnosis codes, outpatient prescription fills, laboratory results, and inpatient admission records. We selected patients who received a nonsteroidal anti-inflammatory drug (NSAID) to predict the development of peptic ulcer disease following the initiation of the drug. The selected cohort comprised $921,167$ samples. 

For the second use case, we considered the task of modeling in-hospital patient mortality. An accurate and timely prediction of this outcome, particularly for patients admitted to an intensive care unit (ICU), can significantly improve quality of care. For this task, we used the Medical Information Mart for Intensive Care (MIMIC III) data~\cite{johnson2016mimic}. MIMIC III is a benchmark data set, from where we derived multivariate time series from over $40,000$ ICU stays and labels to model mortality rate. As discussed in~\cite{harutyunyan2019multitask}, we selected $17$ physiological variables, including demographic details, each comprising $6$ different sample statistic features on $7$ different subsequences of a given time series, resulting in $714$ features per times series. The cohort consisted of $21,139$ ICU stays. 

\subsection{Experimental setup}

For the tasks of ADR and mortality prediction, we used three classification algorithms, amenable to distributed solution using gradient descent, namely perceptron, support vector machine (SVM), and logistic regression. To evaluate the models, prior to and after employing privacy-preserving mechanism, we measure their utility in terms of F1 score. The models were trained on 70\% of the data with 5-fold cross-validation. We considered 10 sites for the federated setup. All experiments were executed on an Intel(R) Xeon(R) E5-2683 v4 2.10 GHz CPU equipped with 16 cores and 64 GB of RAM. The results were reported following 10 rounds of iteration.

\subsection{Comparative analysis}

\begin{figure*}[ht!]
    \centering
	\includegraphics[width=0.41\textwidth]{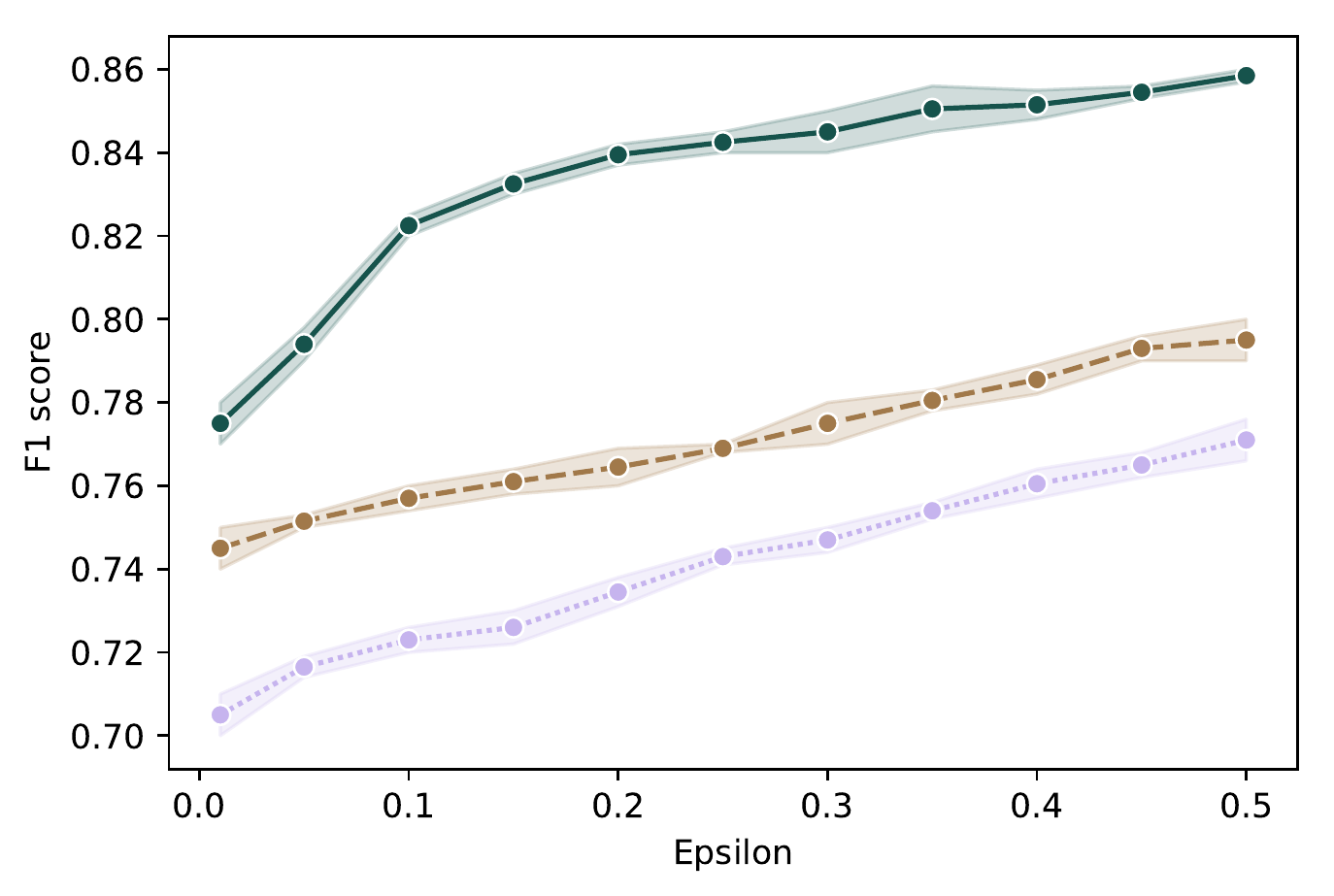} 
	\includegraphics[width=0.565\textwidth]{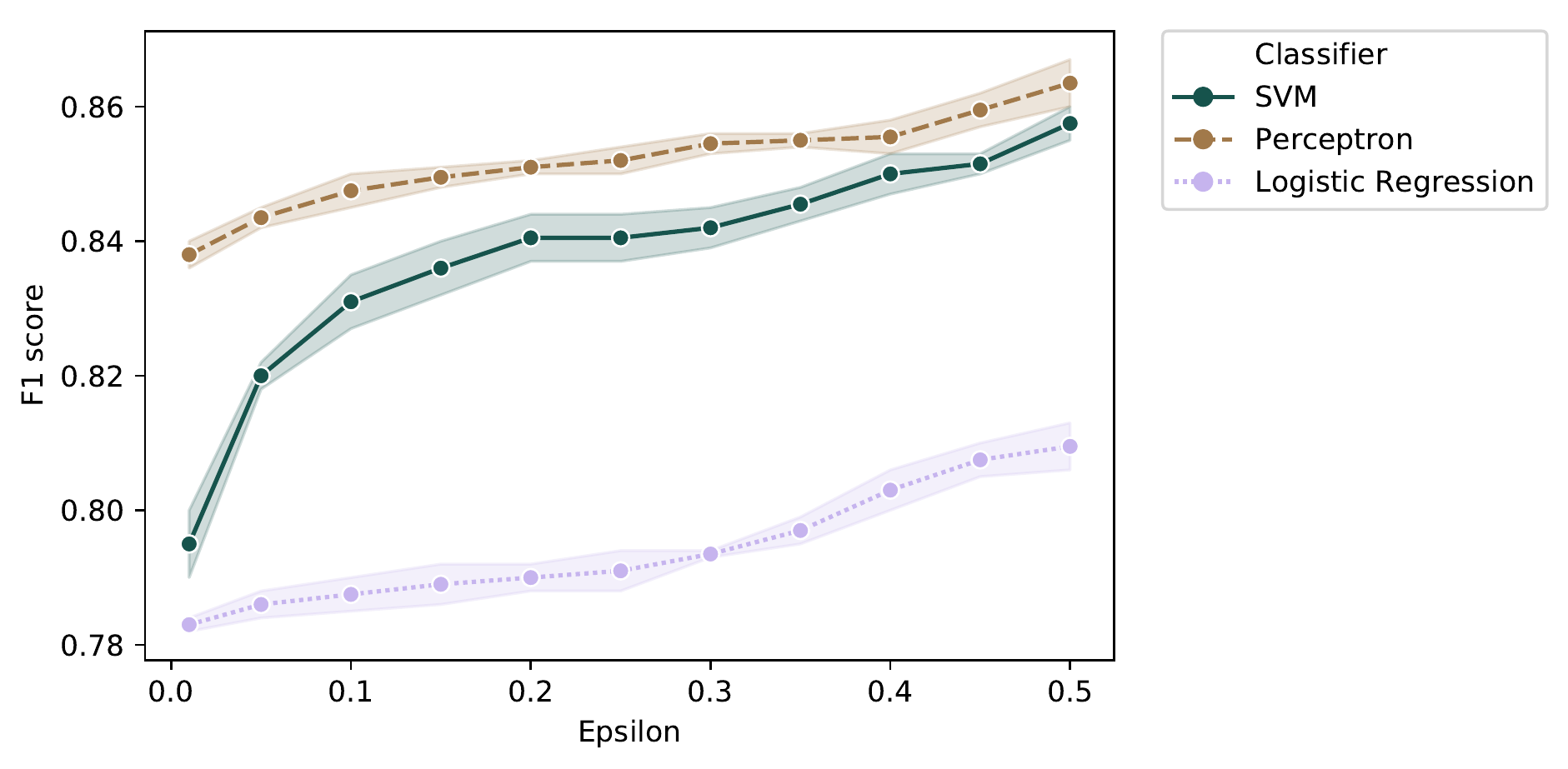}
    \caption{Effect of varying $\epsilon$ in $\epsilon$-differential privacy for (a) ADR prediction using LCED, and (b) in-hospital mortality prediction using MIMIC data.}
    \label{fig:Epsilon_f1_accuracy}
\end{figure*}

To establish benchmark results, we first evaluated the performance of the centralized learning models for the tasks of predicting ADR and in-hospital mortality. We then analyzed the performance of FL models trained on distributed data. For $\epsilon$-differential privacy, we measured the privacy-utility trade-off for a given range of the privacy parameter $\epsilon$. Figure~\ref{fig:Epsilon_f1_accuracy} (a) and (b) present the utility, measured by F1 score, when differential privacy is applied using typical values of parameter $\epsilon\in[0.01, 0.5]$, for LCED and MIMIC data, respectively. As $\epsilon$ increases, the level of privacy degrades, thereby improving the utility or predictive capability of the models. This is consistent across all three classification algorithms and for both datasets. 

\vspace{-.3 cm}

\vspace{-.6 cm}

\begin{figure*}[ht!]
    \centering
    \includegraphics[width=0.42\textwidth]{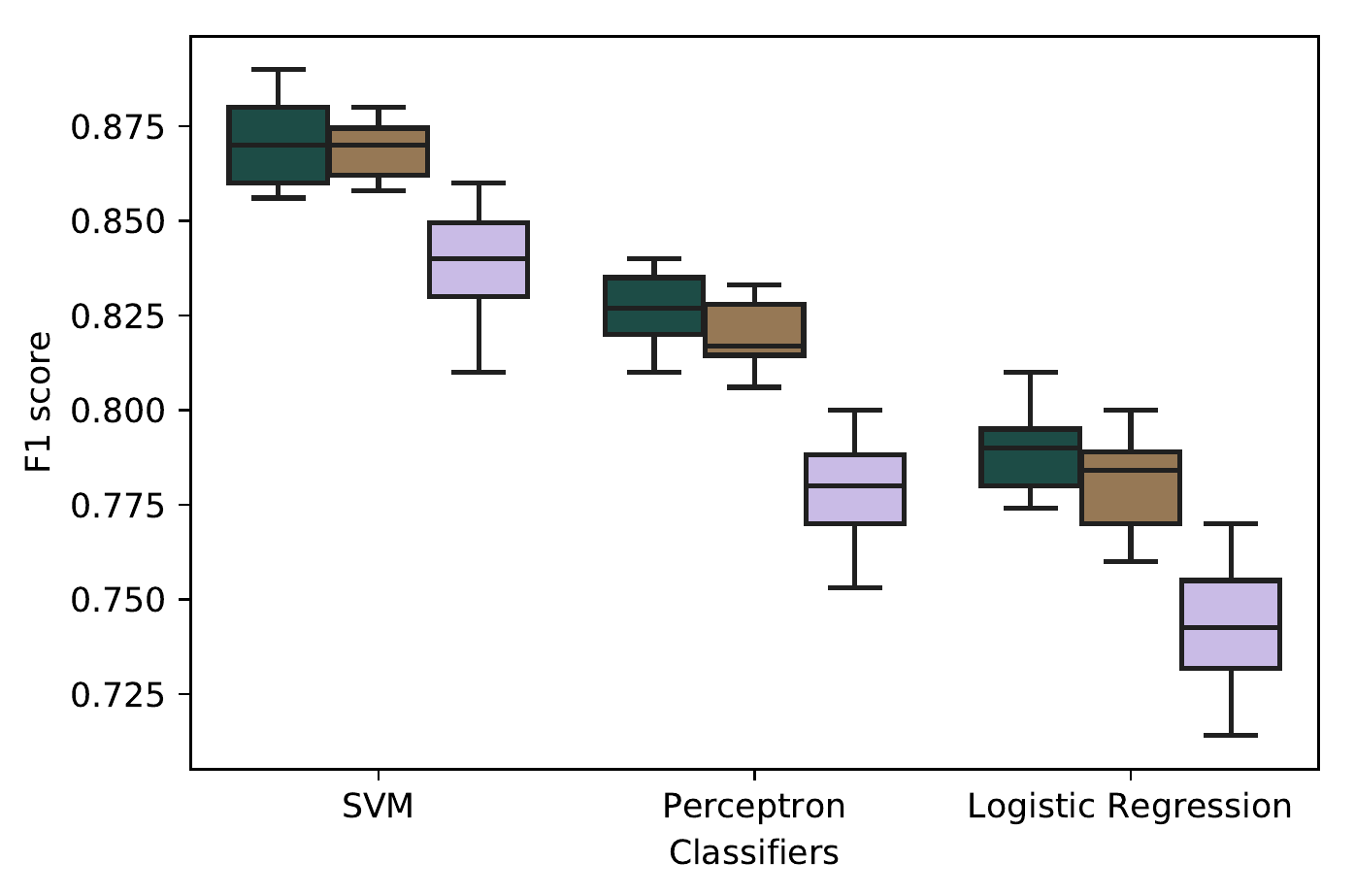}
    \includegraphics[width=0.56\textwidth]{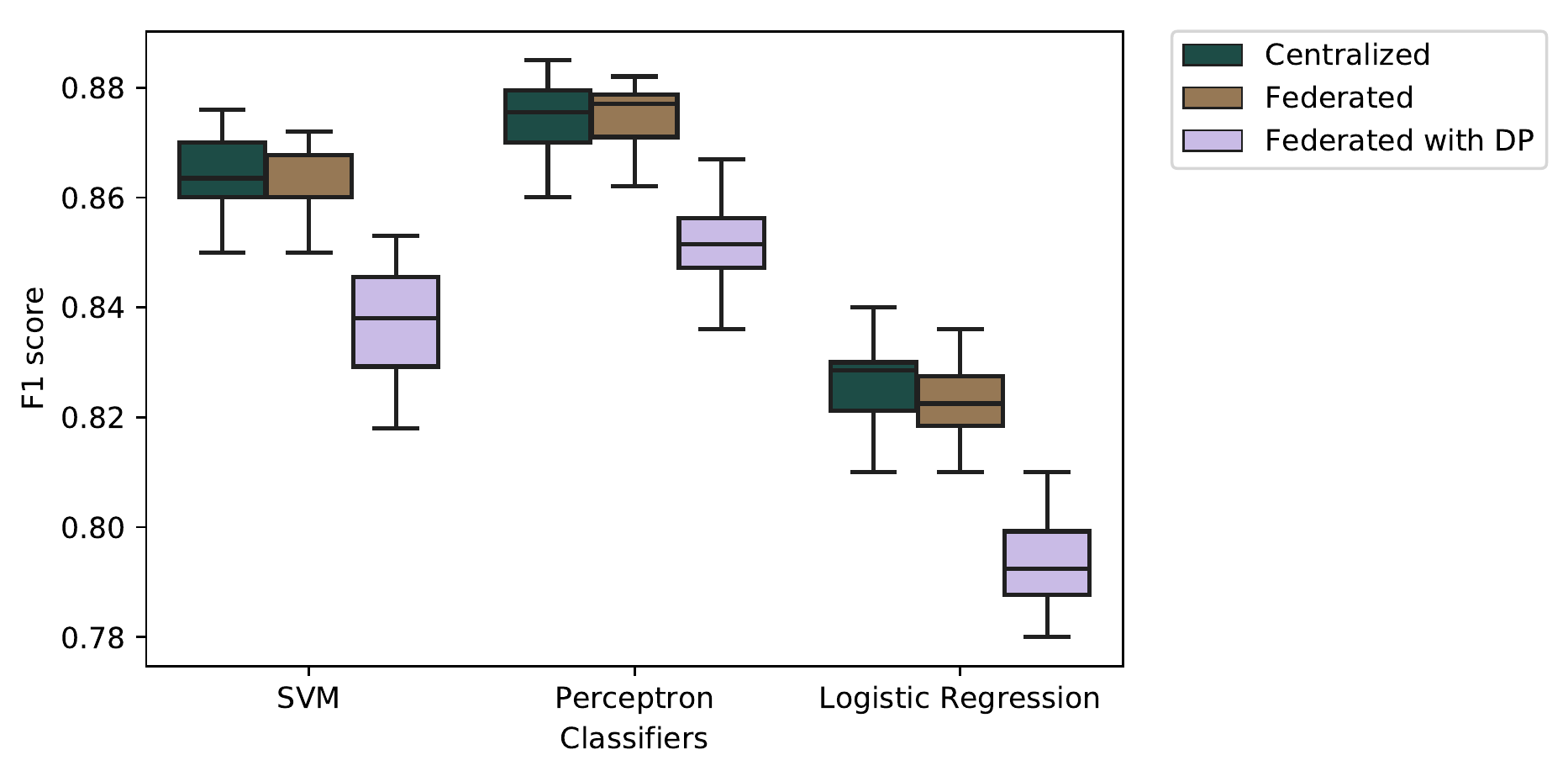} 
    \caption{Comparison of F1 score with (a) LCED, and (b) MIMIC data, between centralized learning, FL, and FL with $\epsilon$-differential privacy (Federated with DP), where $\epsilon = [0.01,0.5]$.}
    \label{fig:compare}
\end{figure*} 

We also compare and contrast the performance of centralized learning, FL, and FL with $\epsilon$-differential privacy in terms of utility, for $\epsilon\in[0.01, 0.5]$. This indicates the level of utility that can be attained for an acceptable range of the privacy parameter $\epsilon$. As shown in Figure~\ref{fig:compare} (a) and (b), FL achieves comparable performance to centralized learning for both tasks of ADR and mortality prediction. Although differential privacy guarantees a given level of privacy, as set by parameter $\epsilon$, it leads to a significant deterioration of the utility of the federated model. 

It must be noted that existing studies advocating the use of differential privacy in a FL setup have not been performed on real data. Moreover, as discussed in~\cite{geyer2017differentially}, the model performance can only be preserved for a very large number of sites, in the order of 1000, but takes a severe hit in the case of fewer sites. Such an assumption of large-scale setup is not realistic for healthcare applications, where sites are typically hospitals or providers, and each site may not have sufficient data for deep learning models to be applicable. This drives the need to leverage data from other sites for constructing more accurate models. Hence, although differential privacy is widely adopted for preserving the privacy of machine learning models, it can yield lower utility in a FL setup, particularly in real-world healthcare applications. This necessitates exploring alternative privacy-preserving approaches that can achieve high model performance while offering sufficient privacy~\cite{choudhury2020anonymizing}.

\section{Conclusion}
The availability of electronic health data poses several opportunities to leverage machine learning for deriving insightful data analytics. In this paper, we implemented a federated learning approach to mitigate the challenges associated with centralized learning. We further explored the potential of differential privacy in preserving the privacy of federated learning models, while maintaining high model performance in the context of real-world health applications. Through experimental evaluation, we show that although differential privacy is being readily adopted in a federated setup, it can lead to a significant loss in model performance for healthcare applications. This necessitates the research and proposal of alternative approaches towards offering privacy in federated learning for healthcare applications.

\bibliography{NIPS_ML4H_arXiv_FedLearn_DifferentialPrivacy_2019}
\bibliographystyle{IEEEtran}

\end{document}